\documentclass[10pt,journal,compsoc]{IEEEtran}
\ifCLASSOPTIONcompsoc
  \usepackage[nocompress]{cite}
\else
  \usepackage{cite}
\fi
\usepackage{graphicx}
\usepackage{float}
\usepackage{multirow}
\usepackage{array}
\usepackage{subcaption}
\usepackage{tabularx}
\usepackage{bm}
\usepackage[export]{adjustbox}
\graphicspath{ {images/} }
\usepackage{amsmath}
\usepackage{amssymb}
\usepackage{cite}
\usepackage{amsmath,amssymb,amsfonts}
\usepackage{graphicx}
\usepackage{textcomp}
\usepackage{xcolor}
\usepackage[]{algorithm2e}
\ifCLASSINFOpdf
\else
\fi
\begin{document}
\title{Quantifying contribution and propagation of error from computational steps, algorithms and hyperparameter choices in image classification pipelines}

\author{Aritra~Chowdhury,
        Malik~Magdon-Ismail, ~\IEEEmembership{Member,~IEEE}
        and~B{\"u}lent~Yener,~\IEEEmembership{Fellow,~IEEE}% <-this % stops a space
\IEEEcompsocitemizethanks{\IEEEcompsocthanksitem A. Chowdhury was with the Department
of Computer Science, Rensselaer Polytechnic Institute, Troy, NY, 12180.\protect\\
E-mail: see http://www.cs.rpi.edu/~achowd/contact.html
\IEEEcompsocthanksitem M. Magdon-Ismail and B. Yener are with Rensselaer Polytechnic Institute}% <-this % stops a space
% \IEEEcompsocthanksitem H.Su is with IBM Research}
% \thanks{Manuscript received July 1, 2018; revised August 26, 2018.}
}

\IEEEtitleabstractindextext{%
\begin{abstract}
Data science relies on pipelines that are organized in the form of interdependent computational steps. Each step consists of various candidate algorithms that maybe used for performing a particular function. Each algorithm consists of  several hyperparameters. Algorithms and hyperparameters must be optimized as a whole to produce the best performance. Typical machine learning pipelines consist of complex algorithms in each of the steps. Not only is the selection process combinatorial, but it is also important to interpret and understand the pipelines. We propose a method to quantify the importance of different components in the pipeline, by computing an error contribution relative to an \textit{agnostic} choice of computational steps, algorithms and hyperparameters. We also propose a methodology to quantify the propagation of error from individual components of the pipeline with the help of a \textit{naive} set of benchmark algorithms not involved in the pipeline. We demonstrate our methodology on image classification pipelines. The \textit{agnostic} and \textit{naive} methodologies quantify the error contribution and propagation respectively from the computational steps, algorithms and hyperparameters in the image classification pipeline. We show that algorithm selection and hyperparameter optimization methods like grid search, random search and Bayesian optimization can be used to quantify the error contribution and propagation, and that random search is able to quantify them more accurately than Bayesian optimization. This methodology can be used by domain experts to understand machine learning and data analysis pipelines in terms of their individual components, which can help in prioritizing different components of the pipeline.
\end{abstract}

\begin{IEEEkeywords}
image classification, hyperparameter optimization, error quantification, algorithm selection
\end{IEEEkeywords}}

\maketitle

% is not selected <OR> if conference mode is selected - because compsoc
% conference papers position the abstract like regular (non-compsoc)
\IEEEdisplaynontitleabstractindextext

% \fi
\IEEEpeerreviewmaketitle

\ifCLASSOPTIONcompsoc
\IEEEraisesectionheading{\section{Introduction}\label{sec:introduction}}
\else
\section{Introduction}
\label{sec:introduction}
\fi
Machine learning and data science have entered many domains of human effort in modern times. The number of self-reported data scientists has doubled in recent years \cite{harrison1995validity}. They have entered various domains including academia, industry and business among others. There has therefore been a demand for machine learning tools that are flexible, powerful and most importantly, interpretable. The effective application of machine learning tools unfortunately requires an expert understanding of the frameworks and algorithms that are present in a machine learning pipeline. It also requires knowledge of the problem domain and understanding of the assumptions used in the analysis. In order for tools to be used adequately by non-experts; new tools must be developed for understanding and interpreting the results of a data analysis pipeline in a specific domain.
\begin{figure}[ht!]
    \centering
    \includegraphics[width=0.5\textwidth]{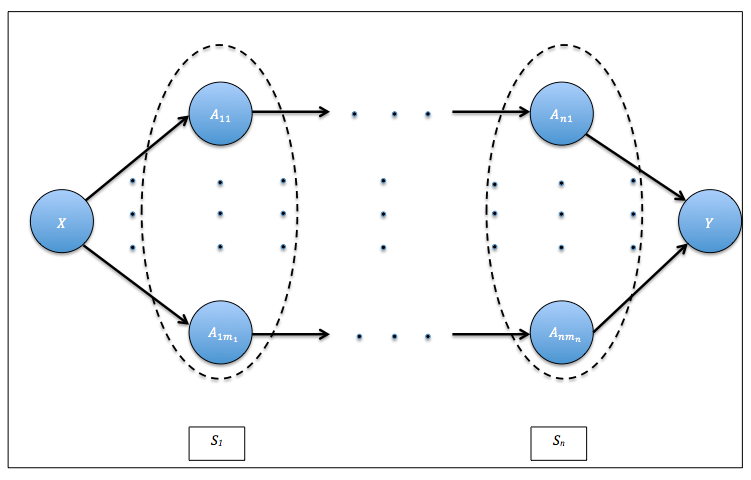}
    \caption{Representation of a data analysis pipeline. This is represented as a generalized directed acyclic graph. $S_i$ represents the $i$-th computational step in the pipeline and $A_{ij}$ represents the $j$-th algorithm in the $i$-th step. $X$ is the input dataset and $Y$ is the evaluation metric.}
    \label{fig:pipeline}
\end{figure}
Pipelines in machine learning and data science are commonly organized in the form of interdependent components. Such components that make up a data analysis pipeline include data preprocessing, feature extraction, feature transformation, model building and model evaluation among others. 
Such pipelines provide a natural way to organize such tasks, and they play a key part in the design and implementation of large scale data science projects. 
% Machine learning toolboxes like scikit-learn \cite{pedregosa2011scikit}, RapidMiner \cite{mierswa2006yale} and Apache Spark \cite{spark2016apache} independently provide frameworks for implementing pipelines. 
% There are frameworks for specialized pipelines such as PARAMO \cite{ng2014paramo} for healthcare data analysis. 

Fig. \ref{fig:pipeline} shows a generic representation of data analysis pipelines as a feed-forward network. Each computational step of the pipeline $S_{i}$ consists of several algorithms ($A_{ij}$) to choose from. Each algorithm in the pipeline has its own hyperparameters $\bm{\theta_{ij}}$ that must be optimized. Therefore, there are an exponential number of combinations of algorithms and hyperparameters in a given data analysis pipeline, which makes it a computationally intensive task to optimize the pipeline. Tuning this pipeline can be viewed as the optimization of an objective function that is noisy and expensive to evaluate. The input to the pipeline is a dataset $X$, the pipeline $P$ (a network consisting of the steps $S_{i}$, the algorithms $A_{ij}$ and corresponding hyperparameters $\bm{\theta_{ij}}$) and the objective to optimize $Y$ such as validation error, accuracy, F1-score, or cross-entropy loss etc. The goal of a data scientist is to find the best set of algorithms and hyperparameters in this pipeline that optimizes the objective function. This corresponds to finding an optimal path through the pipeline in Fig. \ref{fig:pipeline}.  Simple methods such as grid and random search \cite{bergstra2012random} have been used to tackle this problem. More complicated approaches such as Bayesian optimization \cite{snoek2012practical, zhang2016flash} have been used successfully for approaching more difficult problems. Pipeline optimization as a whole has also been approached using genetic algorithms \cite{olson2016evaluation, olson2016tpot}.
We use grid search, random search and Bayesian optimization methods for optimization of the pipeline and each individual path in it. Our present goal is not to improve ways to optimize the pipeline, but to use any one such method to help a domain scientist quantify the importance of different steps in the pipeline. For example "How important is feature extraction?" or "How much of the error is propagated from the \textit{Haralick texture features} algorithm?".

% Interpretation of machine learning pipelines is extremely important for their adoption in various domains. 
% Domain experts prefer to understand how predictive decisions are made by the pipeline. Recently there has been an advent of models and techniques for improving the interpretability of machine learning. \cite{ribeiro2016model} introduces a model-agnostic method for interpreting the results of complex machine learning algorithms. 
% \cite{doshi2017towards} attempts at a definition of interpretability in this context and how it should be measured.
% \cite{koh2017understanding} uses influence functions to understand blackbox predictions. 
In this work, we attempt to provide an analysis of machine learning pipelines in terms of the importance and sensitivity of components in the pipeline (steps, algorithms and hyperparameters) as opposed to the approaches which are geared toward interpretation of algorithms based on the dataset (see \cite{koh2017understanding}). Using our approach, one can understand the importance of different steps like feature extraction and feature transformation and individual algorithms and hyperparameters. To our knowledge, this type of approach to interpretation has not been taken before.
To this end, we propose the understanding of the contribution of error in data analysis pipelines using a method that we denote as the \textit{agnostic} methodology. Essentially, to quantify the contribution of a particular component, we compute the error from the pipeline when the component is selected \textit{agnostically}. In addition, we also propose a \textit{naive} methodology to quantify the error propagated from a particular component down the pipeline. Here, we use a set of benchmark algorithms (that are not a part of the pipeline in question and are in other words \textit{naive} to the pipeline) to quantify the propagated error from the component. We use the cross-entropy loss as the performance metric of the optimization algorithms and basis of error quantification in the image classification pipelines. 
Understanding the importance of the components in the predictive model is important for experts to design better data analysis pipelines. 
Experts can use the information from error contribution and propagation to focus attention on certain parts of the pipeline depending on the source of error. 
% In addition, it also provides non-experts in machine learning insight into the predictions of the model. 
% They can understand which component of the process is more important to in terms of contribution to the final results. 
We introduce a methodology to quantify the contribution of error from different components of the data analysis pipeline, namely the computational steps, algorithms and hyperparameters in the pipeline. 

Pipeline optimization methods and algorithms like grid search, random search \cite{bergstra2012random} and Bayesian optimization \cite{snoek2012practical} are used to optimize the pipeline for performing experiments with our \textit{agnostic} error contribution methodology. We take two different approaches to optimization. The first is hyperparameter optimization (HPO) where a computational path in Fig. \ref{fig:pipeline} is optimized. The second type of optimization is denoted as combined algorithm selection and hyperparameter optimization (CASH). This term was introduced in \cite{thornton2013auto}. This is a more difficult problem, because the pipeline is optimized globally, in that the result of the optimization is a single optimized path that produces the best performance over all the paths in the machine learning workflow.  

We use four datasets to demonstrate the error quantification methodology. The problem we focus on is image classification. We show the performance of the optimization frameworks (HPO and CASH) for the experiments. We show experimentally that CASH using random search and Bayesian optimization can be efficiently used for quantification of errors from the different computational steps of the pipeline. In addition, HPO frameworks of both Bayesian optimization and random search provides  estimates of error quantification from the algorithms and hyperparameters in a particular path of the pipeline.
We demonstrate from the results that the \textit{agnostic} error contribution and the \textit{naive} error propagation methodologies maybe used by both data science and domain experts to improve and interpret the results of image classification pipelines. In addition, we observe that random search is a more accurate estimator of error quantification than Bayesian optimization. 

% The paper is organized as follows. Section \ref{sec2} describes the methods that are used in this work. In section \ref{sec3}, we propose the methodologies of error contribution and propagation. The experimental frameworks, datasets, results and discussion makes up section \ref{sec4}. This is followed by the description an application of the work in section \ref{sec:application} and the conclusion in section \ref{sec5}.

\section{Foundations}
\label{sec2}
In this section we describe the optimization problem and methods that are used in this work. 
\subsection{Algorithm selection and hyperparameter optimization}
\label{subsec_AS_HPO}
We approach the problem of optimization of the pipeline from the following frameworks. 
% In one framework, each path in the pipeline in Fig. \ref{fig:pipeline} is individually optimized. This essentially boils down to the problem of hyperparameter optimization (HPO)  because the hyperparameters of each algorithm are optimized for each individual path. In the second framework, the entire pipeline is optimized. This means that the algorithms and hyperparameters are optimized together. This is denoted as combined algorithm selection and hyperparameter optimization (CASH). 

\subsubsection{Hyperparameter optimization (HPO)}
\label{subsubsec_HPO}
Let the \textit{n} hyperparameters in a path be denoted as $\theta_1, \theta_2, ..., \theta_n$, and let $\Theta_1, \Theta_2, ..., \Theta_n$ be their respective domains. The hyperparameter space of the path is  \textbf{$\Theta$} = $\Theta_1 \times \Theta_2 \times ... \times \Theta_n$.

When trained with $\emph{$\theta$} \in \textbf{$\Theta$}$ on data $D_{train}$, the validation error is denoted as \par
\begin{figure}[ht!]
\begin{adjustbox}{right}
  \begin{subfigure}{\columnwidth}
      \includegraphics[width=\linewidth]{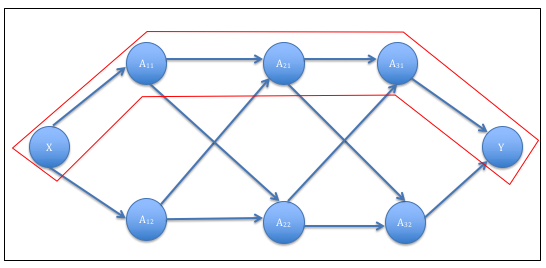}
      \caption{hyperparameter optimization in a data analysis pipeline. Each path in the pipeline is individually optimized.}
      \label{fig:HPO}
  \end{subfigure}
\end{adjustbox}

\begin{adjustbox}{right}
  \begin{subfigure}{\columnwidth}
      \includegraphics[width=\linewidth]{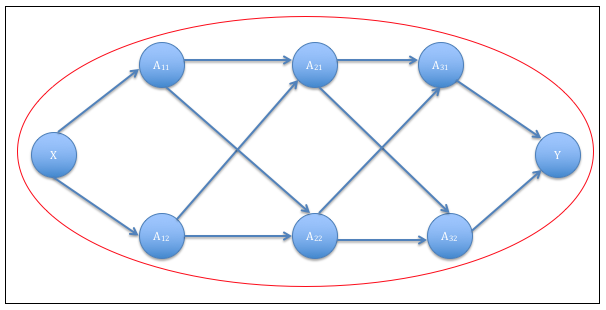}
      \caption{Combined algorithm selection and hyperparameter optimization (CASH) framework. The entire pipeline is optimized simultaneously.}
      \label{fig:CASH}
  \end{subfigure}
\end{adjustbox}
\caption{Optimization frameworks}\label{fig:frameworks}
\end{figure}

\noindent $\mathcal{L}(\theta, D_{train}, D_{valid})$. Using $k$-fold cross-validation, the hyperparameter optimization problem for a dataset $D$ is to minimize:
\begin{equation}
f^D(\theta) = \frac{1}{k}\sum_{i=1}^{k} \mathcal{L}(\emph{$\theta$}, D_{train}^{(i)}, D_{valid}^{(i)})
\label{eq:hpo}
\end{equation}
Hyperparameters $\theta$ may be numerical, categorical or conditional with a finite domain. The minimization of this objective function provides the optimal configuration of hyperparameters on a particular path in the pipeline in Fig. \ref{fig:pipeline}. The optimization of the objective function defined by Eq. \ref{eq:hpo} is very expensive. Depending on the type of hyperparameter variables, the derivatives and convexity properties maybe unknown, and derivative free global optimization methods like Bayesian optimization and techniques like random search maybe used to tackle this problem. This framework is represented in Fig. \ref{fig:HPO}.

% \begin{figure}[ht!]
%     \centering
%     \includegraphics[width=0.5\textwidth]{HPO}
%     \caption{Hyperparameter optimization in a data analysis pipeline. Each path in the pipeline is individually optimized.}
%     \label{fig:HPO}
% \end{figure}

\subsubsection{Combined algorithm selection and hyperparameter optimization (CASH)}
\label{subsubsec_CASH}
We can define the CASH formulation using Fig. 1. Let there be $n$ computational steps in the pipeline. Each step $i$ in the pipeline consists of algorithms $A_i(\Theta_i)$, where $A_i(\Theta_i) = \{A_{i1}(\theta_{i1}), ..., A_{im_{i}}(\theta_{im_{i}})\}$, $m_{i}$ is the number of algorithms in step $i$, $A_{ij}$ represents the $j$-th algorithm in step $i$, and \textbf{$\theta_{ij}$} represents the set of hyperparameters corresponding to  $A_{ij}$. The entire space of algorithms and hyperparameters is therefore given by \par
\noindent $\mathcal{A} = A_1(\Theta_1) \times A_2(\Theta_2) \times ... \times A_n(\Theta_n)$. The objective function to be minimized for CASH is given by
\begin{equation}
f^D(A) = \frac{1}{k}\sum_{i=1}^{k} \mathcal{L}(\emph{A}, D_{train}^{(i)}, D_{valid}^{(i)}) 
\label{eq:cash}
\end{equation}
where, $A \in \mathcal{A}$ and other notations are the same as those introduced in the previous section.
Similar to the objective function defined over the hyperparameters in Eq. \ref{eq:hpo}, the optimization in Eq. \ref{eq:cash} is even more difficult due to the additional problem of algorithm selection. Again, the derivates may be impossible to compute and convexity properties may be completely unknown. This framework is represented in Fig. \ref{fig:CASH}.

% \begin{figure}[ht!]
%     \centering
%     \includegraphics[width=0.5\textwidth]{CASH}
%     \caption{Combined algorithm selection and hyperparameter optimization (CASH) framework. The entire pipeline is optimized simultaneously.}
%     \label{fig:CASH}
% \end{figure}

\subsection{Optimization methods}
\label{subsec:optimization}
The critical step in HPO or CASH is to choose the set of trials in the search space, which is $\Theta$ for HPO and $\mathcal{A}$ for CASH. Various algorithms and frameworks \cite{li2016hyperband, feurer2015initializing, maclaurin2015gradient, bergstra2013making} have been developed for both these frameworks. 
In this section, methods that are used in this paper for optimization of Eq. \ref{eq:hpo} and Eq. \ref{eq:cash} are described. Grid search, random search and Bayesian optimization are used in this work.
\subsubsection{Grid search}
\label{grid}
Grid search is the simplest and the most widely used of all methods for coming up with trials in the search space. The set of trials in grid search is formed by assembling every possible set of values in $\Theta$ (HPO) and $\mathcal{A}$ (CASH) and computing the validation loss for each. The configuration $\theta \in \Theta$ or $A \in \mathcal{A}$ that minimizes the validation loss $\mathcal{L}$ is chosen as the optimum configuration. Unfortunately, grid search is computationally very expensive. For HPO, the number of trials corresponds to $\prod_{i=1}^n |\Theta_i|$, and for CASH this is $\prod_{i=1}^n |A_i(\Theta_i)|$. This product makes grid search suffer from the \textit{curse of dimensionality}. This is because the number of trials grows exponentially with the number of hyperparameters. However, grid search has certain advantages. Firstly, parallelization and implementation is trivial. It is reliable in low dimensional spaces. In addition, grid search is robust in the sense that results maybe replicated easily. 

\subsubsection{Random search}
\label{random}
Random search is the optimization method where trial configurations are randomly sampled from the search space of $\Theta$ (HPO) or $\mathcal{A}$ (CASH). \cite{bergstra2012random} shows empirically and theoretically that randomly selecting trials is sufficiently accurate and more efficient than performing optimization using grid search. It is actually more practical than grid search because it can be used with a cluster of computers. This is because in case of failure of some of the nodes, the experimenter can change the resolution of the search by adding new trials or ignoring failed trials completely. The trials are i.i.d unlike grid search. Random search is also more efficient than grid search in higher dimensional spaces as shown in \cite{bergstra2012random}.
% It has all the practical advantages that grid search offers (simplicity, ease of implementation, trivial parallelism) while trading a small reduction of efficiency in low dimensional spaces with an improvement of computational efficiency in large dimensional spaces.
We show that random search is the better than Bayesian optimization in quantifying the error contribution and propagation in machine learning pipelines.

\subsubsection{Bayesian optimization}
\label{bayesian}
Sequential model based Bayesian optimization (SMBO) \cite{hutter2011sequential} is the method of choice when it comes to optimization of complicated black-box functions. In a nutshell, it consists of two components. The first is a probabilistic model and the second is an acquisition function. The probabilistic model can be modelled using Gaussian processes (Spearmint) \cite{snoek2012practical}, random forests (SMAC) \cite{hutter2011sequential} and using density estimation with Tree-structured Parzen estimators (TPE) \cite{bergstra2011algorithms}. The acquisition function determines the future candidates or trials for evaluation. The acquisition function is relatively cheap to evaluate compared to the actual objective function $f^D$. One of the most prominent acquisition functions is \textit{expected improvement} (EI) \cite{expected_improvement}. We use the sequential model-based algorithm configuration (SMAC) that uses random forests as the Bayesian optimization framework. This is because it can be used for optimizing conditional hyperparameter configurations. The choice is also based on empirical results in \cite{eggensperger2013towards}.

\section{Proposed methods}
\label{sec3}
In this section the proposed methodologies for quantification of error contribution and propagation are presented. The methods are independent of the optimization methods that maybe used for both the HPO and CASH formulations.

\subsection{Error contribution with the \textit{agnostic} methodology}
\label{EQ}
% Machine learning pipelines maybe understood and interpreted by quantifying the contribution of error from different parts of the pipeline. For example, it is useful for machine learning experts and domain experts to understand and identify where the source of the error is in a pipeline. Referring back to Fig. \ref{fig:pipeline}, if it was known that most of the error in the final result originated from feature extraction, then machine learning practitioners would devote more time and energy to coming up with better algorithms for feature extraction or would try to fine-tune the algorithms in that step to reduce the error. 
% In addition, if it were possible to quantify the contribution of errors from certain algorithms or hyperparameters in the pipeline, then the data scientists would try to fine tune the algorithms in the pipeline or even try to replace the algorithms with better alternatives. 

We propose an \textit{agnostic} methodology for quantifying error contributions from different parts of the pipeline. It is defined as the error obtained by being agnostic to a particular component of the pipeline (computational step, algorithms or hyperparameters). This refers to randomly picking the sub-components of that component while optimizing the rest of the pipeline. We shall define what \textit{agnostic} means for computational steps, algorithms and hyperparameters individually.

\subsubsection{Error contribution from computational steps}
\label{subsubsec_eq_steps}
% The \textit{agnostic} methodology maybe used for quantification of contributions from computational steps like feature extraction, data pre-processing and learning algorithms. 
Being \textit{agnostic} to a computational step means that the algorithms in that step are selected randomly for that step while the remaining pipeline is optimized. The average of the minimum errors obtained with each algorithm in the step used as the only algorithm in that particular step, provides an estimate of the agnostic error from a particular pipeline.  
More formally, the agnostic methodology is defined for computational steps in the following manner. Using Fig. \ref{fig:pipeline} as a reference, let $n$ be the number of steps in the pipeline. Each step in the pipeline is denoted as $S_i$. $|S_i|$ is the number of algorithms in step $i$. $A_{ij}$ denotes the $j$-th algorithm in the $i$-th step. $E_{A_{ij}}^*$ is the minimum  validation error found with $A_{ij}$ as the only algorithm in step $i$. $E^*$ represents the minimum validation error found after optimization of the entire pipeline (using the CASH framework).  The error contribution from step $i$, $EC_{S_i}^*$ is given by Eq. \ref{eq:eq_step}.
\begin{equation}
\label{eq:eq_step}
EC_{S_i}^* = \frac{1}{|S_i|}\sum_{j=1}^{|S_i|} E_{A_{ij}}^* - E^*,
\end{equation}
where, $i = {1, ..., n}, j = {1, ..., |S_i|}$
Taking the difference with respect to the global minimum in Eq. \ref{eq:eq_step} provides an estimate of the error contribution from step $i$ of the pipeline. A large value of $EC_{S_i}^*$ would mean that step $S_i$ is important for the pipeline. Therefore, more attention should be paid to optimizing that step of the pipeline or improving the algorithms that are used for that step.

\subsubsection{Error contribution from algorithms}
\label{subsubsec_eq_alg}
Similar to the \textit{agnostic} methodology for steps, we define the \textit{agnostic} methodology for algorithms. In this case, we focus on a single path in the pipeline in Fig. \ref{fig:pipeline} optimized using the HPO framework in Fig. \ref{fig:HPO}. 
% We optimize a path instead of the entire pipeline because it does not make sense to compute the error contribution of algorithms or hyperparameters with respect to the pipeline. This is because the global minimum error $E^*$ of the entire pipeline may lie on a completely different path that does not involve algorithm $A_{ij}$. Therefore, a comparison of the \textit{agnostic} error of algorithms or hyperparameters with respect to the global minimum error would not make sense.

Let's assume we are trying to quantify the error contribution of a particular algorithm $A_{ij}$ that lies on path $p$. Being $agnostic$ to $A_{ij}$ means we optimize everything else on the path except the algorithm. This means that we pick the hyperparameters $\bm{\theta_{ij}}$ of algorithm $A_{ij}$ randomly while optimizing the rest of the algorithms on the path. This is formally calculated by taking the average of the optimum errors on the path for each configuration of $\bm{\theta_{ij}}$. The minimum validation error on the path is then subtracted from this error to give us the error contribution from algorithm $A_{ij}$ on path $p$. These errors are computed using the results and the search trials on the HPO framework in section \ref{subsubsec_HPO}.

\begin{equation}
\label{eq:eq_alg}
EC_{A_{ij}^p}^* = \frac{1}{|\bm{\theta_{ij}}|}\sum_{z=1}^{|\bm{\theta_{ij}}|} {E_{A_{ij}}^z}^* - {E_{A_{ij}^p}^*},
\end{equation}

where, $i = {1, ..., n}, z = {1, ..., |\bm{\theta_{ij}}|}$, $|\bm{\theta_{ij}}|$ represents the number of hyperparametric configurations of $A_{ij}$,  ${E_{A_{ij}}^z}^*$ is the minimum error obtained with the $z$-th configuration of $\bm{\theta_{ij}}$ and $E_{A_{ij}^p}^*$ is the minimum error found over the path $p$ that consists of algorithm $A_{ij}$. $EC_{A_{ij}^p}^*$ is the error contribution of algorithm $A_{ij}$ over the path $p$ in the pipeline.

\subsubsection{Error contribution from hyperparameters}
\label{subsubsec_eq_hyper}
In the case of hyperparameters, again we focus on a single path similar to what we did for algorithms. Let's assume we are trying to quantify the error contribution of a particular hyperparameter $\bm{\theta_{ijk}}$ that lies on path $p$, i.e. the $k$-th hyperparameter of the $j$-th algorithm in the $i$-th step of the pipeline. Being $agnostic$ to $\bm{\theta_{ijk}}$ means we optimize everything else on the path except the hyperparameter. This means that we pick the hyperparameter $\bm{\theta_{ijk}}$ of algorithm $A_{ij}$ randomly while optimizing the rest of the hyperparameters on the path. This is formally calculated by taking the average of the optimum errors on the path for each configuration of $\bm{\theta_{ijk}}$. The minimum validation error on the path is then subtracted from this error to give us the error contribution from hyperparameter $\bm{\theta_{ijk}}$ on path $p$. This is again computed using the HPO framework described in section \ref{subsubsec_HPO}.

\begin{equation}
\label{eq:eq_hyper}
EC_{\bm{\theta_{ijk}}^p}^* = \frac{1}{|\bm{\theta_{ijk}}|}\sum_{z=1}^{|\bm{\theta_{ijk}}|} {E_{\bm{\theta_{ijk}}}^z}^* - {E_{A_{ij}^p}^*},
\end{equation}

where, $i = {1, ..., n}, z = {1, ..., |\bm{\theta_{ij}}|}$, $k$ = number of hyperparameters of algorithm $A_{ij}$. $|\bm{\theta_{ijk}}|$ represents the number of configurations of $\bm{\theta_{ijk}}$,  ${E_{\bm{\theta_{ijk}}}^z}^*$ is the minimum error obtained with the $z$-th configuration of $\bm{\theta_{ijk}}$ and $E_{A_{ij}^p}^*$ is the minimum error found over the path $p$ that consists of algorithm $A_{ij}$. $EC_{\bm{\theta_{ijk}}^p}^*$ is the error contribution from hyperparameter $\bm{\theta_{ijk}}$ of algorithm $A_{ij}$ that lies on path $p$ of the pipeline.

\subsection{Error propagation using the \textit{naive} methodology}
\label{subsec:naive}
\begin{figure}[ht!]
\label{naive_pipeline}
    \centering
    \includegraphics[scale=0.33]{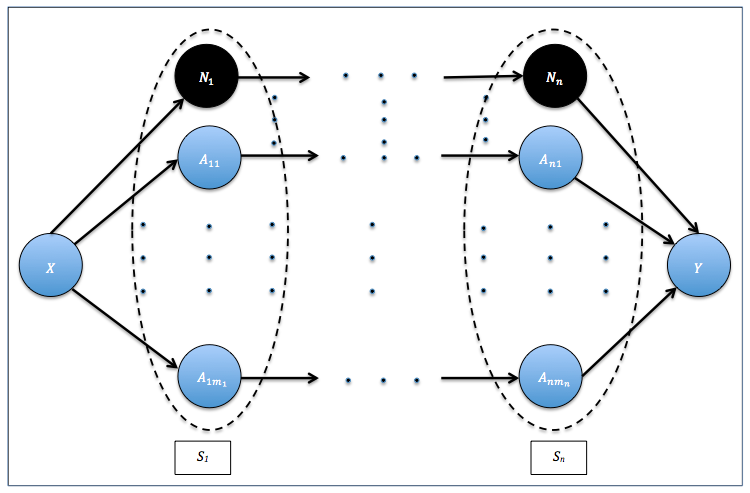}
    \caption{Naive methodology used in the error propagation model. A \textit{naive} algorithm $N_i$ is added to each step the general pipeline in Fig. \ref{fig:pipeline}. The naive algorithms are used as a benchmark for the error propagation model.}
    \label{fig:naive}
\end{figure}

Understanding how error propagates along a machine learning pipeline is important for it's analysis. The final error contribution that is obtained in a pipeline (from each component) consists of two sources. The first is the actual error that originates from the various components of the pipeline. The second is the error that is accumulated along the pipeline because of the propagated error. 
% For example an algorithm like principal components analysis that is used for transforming the features may provide a better performance with one feature extraction algorithm over another. Data scientists and domain experts alike may use this tool to understand and analyze a pipeline in detail to make sense of how the error propagates. 
We propose a method to model the error propagation for steps, algorithms and hyperparameters using the error contributions computed using the formulation in the previous section. This is done by introducing \textit{naive} algorithms at each step of the pipeline. This model is represented in Fig. \ref{fig:naive}. 
As shown in the figure above, \textit{naive} algorithms are introduced at each step. They are denoted as $N_i$. This naive formulation is an integral component for the error propagation model. The \textit{naive} algorithms are not a part of the pipeline. They are used as a benchmark for building the error propagation model that we propose in this section. We define 6 different types of errors using this approach. These errors are represented in Table \ref{table:naive_errors}.

\begin{table}[h!]
\centering
\caption{Notations and definitions of error values used in the development of the error propagation model}
\begin{tabularx}{\linewidth}{ |X|X| }
 \hline
 Notation & Definition \\ 
 \hline
 $E_{opt->opt}$ & Minimum error found by using the optimum algorithm or hyperparameter in the current step or algorithm respectively and optimizing all the steps or algorithms that proceed it in the pipeline. \\ 
 \hline
 $E_{agnostic->opt}$ & Minimum error found by using the $agnostic$ methodology for the current step or algorithm and optimizing all the steps or algorithms that proceed it in the pipeline. \\ 
 \hline
  $E_{naive->opt}$ & Minimum error found by using the naive algorithm for the current step or algorithm and optimizing all the steps or algorithms that proceed it in the pipeline. \\ 
 \hline
 $E_{opt->naive}$ & Minimum error found by using the optimum algorithm or hyperparameters for the current step or algorithms respectively and naive algorithms for all the steps or algorithms that proceed it in the pipeline. \\ 
 \hline
 $E_{naive->naive}$ & Minimum error found by using the naive algorithm for the current step or algorithm and naive algorithms for all the steps that proceed it in the pipeline. \\ 
 \hline
 $E_{agnostic->naive}$ & Minimum error found by using the $agnostic$ methodology for the current step or algorithm and naive algorithms all the steps that proceed it in the pipeline.\\ 
 \hline
 \end{tabularx}
\label{table:naive_errors}
\end{table}

Let us now define 3 types of error differences using the notations in Table 1.
\begin{equation}
\label{eq:naive_errors}
\begin{aligned}
\Delta E_1 =& E_{agnostic->opt} - E_{opt->opt} \\
\Delta E_2 =& E_{agnostic->naive} - E_{opt->naive} \\ 
\Delta E_3 =& E_{naive->naive} - E_{naive->opt}
\end{aligned}
\end{equation}
 $\Delta E_1$ is the same as the error contribution $EC_{S_i}^*$ for steps, $EC_{A_{ij}^p}^*$ for algorithms and $EC_{\bm{\theta_{ijk}}^p}^*$ for hyperparameters described in Section \ref{EQ}. It is the error contribution from a particular step, algorithm or hyperparameter with respect to the pipeline or path. $\Delta E_2$ is the error contribution from a component with naive algorithms in the steps proceeding it. This value represents the error contribution when, a set of benchmark (\textit{naive}) components are selected instead of the optimal components. $\Delta E_3$ is the quantification of the difference of error propagated over the \textit{naive} algorithms as opposed to propagating the error over the optimal set of algorithms.  Let us define the actual or direct error contributed by a component as $\alpha$ and the propagated error as a multiple of this error (this is an assumption of the model). Let this be denoted as $\gamma\alpha$. We define $\gamma$ as the \textit{propagation factor}. $\Delta E_1$ can be be broken down into the sum of the direct error $\alpha$ and the propagated error $\gamma\alpha$. Then we can write $\Delta E_1$ as,
\begin{equation}
\Delta E_1 = \alpha + \gamma\alpha
\label{eq:E1}
\end{equation}
$\Delta E_2$ consists of the error from $\Delta E_1$ and some additional propagated error. This is because the first components on the right hand side of the $\Delta E_1$ and $\Delta E_2$ are the same as we can see from Eq. \ref{eq:naive_errors}. The additional error in $\Delta E_2$ is due to the error being propagated over the \textit{naive} algorithms instead of the optimal algorithms. This additional propagated error is quantified by $\Delta E_3$. Assuming that the propagation factor remains constant over the pipeline, the additional propagated error is given by $\gamma\Delta E_3$. Therefore, we can formalize this as:
\begin{equation}
\Delta E_2 = \alpha + \gamma(\alpha + \Delta E_3)
\label{eq:E2}
\end{equation}
Let us denote the actual error as $E_{direct}$ which is the same as $\alpha$, and the propagated error as $E_{propagation}$ which is $\gamma\alpha$. Solving Eq. \ref{eq:E1} and Eq. \ref{eq:E2} simultaneously, we get the values for $E_{direct}$, $E_{propagation}$ and the \textit{propagation factor} $\gamma$ as,
\begin{equation}
\begin{aligned}
E_{direct} =& \frac{\Delta E_1 \Delta E_3}{\Delta E_2 + \Delta E_3 - \Delta E_1} \\
E_{propagation} =& \frac{\Delta E_1 (\Delta E2 - \Delta E_1)}{\Delta E_2 + \Delta E_3 - \Delta E_1}\\
\gamma =& \frac{\Delta E_2 - \Delta E_1}{\Delta E_3}
\end{aligned}
\label{eq:EP}
\end{equation}
We use this model of error propagation to quantify the direct error ($E_{direct}$)the propagated error ($E_{propagated}$) and the propagation factor ($\gamma$). These values maybe easily computed by running the optimization methods described in section \ref{subsec:optimization} over the optimization frameworks explained in section \ref{subsec_AS_HPO}.
We show the results of the experiments using this model in the following section.

\section{Experiments and results}
\label{sec4}
\begin{figure}[ht!]
    \centering
    \includegraphics[scale=0.3]{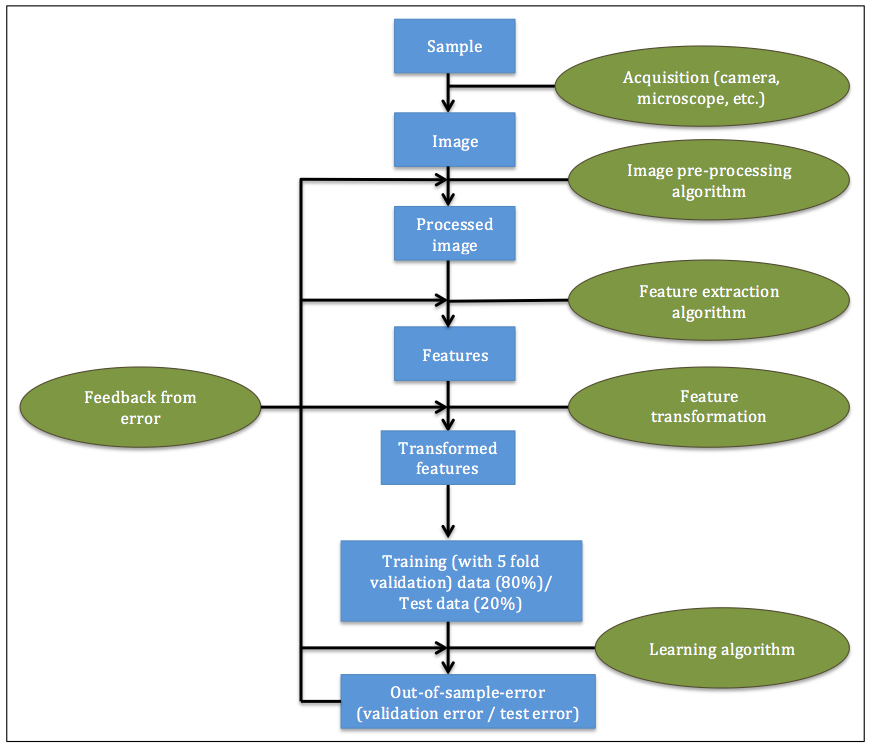}
    \caption{Representation of an image classification workflow. The pipeline consists of the steps represented by green ellipses and the outputs of each step represented by blue rectangles. In this work, we focus on the steps and outputs after pre-processing.}
    \label{fig:flowchart}
\end{figure}

\begin{figure}[ht!]
    \centering
    \includegraphics[scale=0.45]{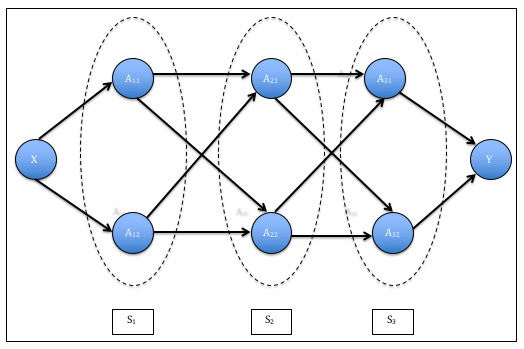}
    \caption{Representation of the image classification pipeline as a feedforward network. This is an instantiation of the generalized data analysis pipeline in Fig. \ref{fig:pipeline}}
    \label{fig:images_pipeline}
\end{figure}

In this section, we describe the experiments performed on the data analysis pipeline to quantify the error contribution and propagation from different components. Image classification is the data analysis problem chosen for demonstrating the error quantification experiments. A representation of an image classification pipeline is shown in Fig. \ref{fig:flowchart} in the form of a flowchart.  
In this work, we focus on real world scientific datasets from the domains of medical pathology and material science.  Therefore, the flowchart starts with a sample that is imaged with acquisition technology like a camera or a microscope. The image is then processed using image pre-processing algorithms like normalization and standardization. This may also include image segmentation algorithms. This is followed by feature extraction algorithms that extract useful information from the images. Sometimes, the features extracted are transformed to a different vector space using feature transformation (feature selection or dimensionality reduction) algorithms. The dataset is then divided into training and test datasets in a 80-20 split. Finally, classification algorithms like Random forests \cite{breiman2001random} and support vector machines (SVM) \cite{cortes1995support} are used for learning in order to build a predictive model for the image classification problem. The performance of the pipeline is evaluated using classification metrics like F1-score, accuracy, precision and recall. We use the cross entropy loss on the validation data as the estimate of the out-of-sample error. This error is then used as a feedback to quantify the contribution and propagation of the errors from different components of the pipeline. The specific pipeline used in this work is shown in Figure \ref{fig:images_pipeline}.

There are 3 computational steps in this pipeline, namely feature extraction ($S_1$), feature transformation ($S_2$) and learning algorithms ($S_3$). The steps, algorithms and corresponding hyperparameters $A_{ij}(\theta_{ij})$ are described in Table \ref{table:algorithms_table}.

\begin{table}[ht!]
\centering
\caption{Algorithms and hyperparameters used in the image classification pipeline. The specific algorithms and corresponding \textit{hyperparameters} are defined in the last column}
\begin{tabularx}{\linewidth}{ |X|X|X| } 
 \hline
 Step & $A_{ij}(\theta_{ij})$ & Definition \\ 
 \hline
 \multirow{3}{*}{Feature extraction} & $A_{11}(\theta_{11})$ & Haralick texture features (\textit{Haralick distance}) \\ 
 & $A_{12}(\theta_{12})$ & Pre-trained CNN trained on ImageNet \cite{deng2009imagenet} database with VGG16 \cite{simonyan2014very} network  \\
%   & $A_{13}(\theta_{13})$ & Pre-trained CNN trained on ImageNet \cite{deng2009imagenet} database with Inception \cite{szegedy2016rethinking} network  \\
 \hline
 \multirow{3}{*}{Feature} & $A_{21}(\theta_{21})$ & PCA (\textit{Whitening}) \cite{wold1987principal} \\
 {transformation} & $A_{22}(\theta_{22})$ & ISOMAP (\textit{Number of neighbors, Number of components}) \cite{tenenbaum2000global} \\
 \hline
 \multirow{3}{*}{Learning algorithms} & $A_{31}(\theta_{31})$ & Random forests (\textit{Number of estimators, Maximum features}) \cite{breiman2001random} \\
 & $A_{32}(\theta_{32})$ & SVM ($C, \gamma$) \cite{cortes1995support}\\
 \hline
 \end{tabularx}
 \label{table:algorithms_table}
\end{table}
The algorithms described in Table \ref{table:algorithms_table} are selected for making up the components of the pipeline in Fig. \ref{fig:images_pipeline}. This pipeline is meant to serve as an example for demonstrating the experiments using the error contribution framework described in section \ref{sec3}. It can easily be generalized to any data analysis problem that involve pipelines.

\subsection{Optimization frameworks}
\label{frameworks}
Experiments are performed using two optimization frameworks. These frameworks have been described in detail in Section \ref{subsec_AS_HPO}. 
The first global optimization framework is the CASH framework described in Section \ref{subsubsec_CASH}. It is depicted in Fig. \ref{fig:CASH}. Here, the pipeline is optimized as a whole including the algorithms, which are themselves considered as hyperparameters in this framework. This is used for quantification of the contribution of error with respect to \textit{computational steps} in the pipeline.

The second framework shown in Fig. \ref{fig:HPO} is the hyperparameter optimization (HPO) framework where each path in the pipeline is optimized individually. This is described in detail in section \ref{subsubsec_HPO}. This framework is used for quantifying the contribution and propagation of error with respect to \textit{algorithms} and \textit{hyperparameters} in the each path of the pipeline.
Specifically, we choose the path \textit{Haralick texture features} - \textit{ISOMAP} - \textit{Random forests} to demonstrate the error quantification approach for algorithms. This path is chosen to demonstrate the error contribution and propagation methodology because this path consists of the most number of hyperparameters. The methodologies can however be run on any path of the pipeline.

\subsection{Datasets}
Four datasets from the domains of medicine and material science are used in this work. They are image datasets of breast cancer\cite{bilgin2007cell}, brain cancer \cite{gunduz2004cell}, and two datasets of microstructures in material science \cite{chowdhury2016image}. They are described in Table \ref{table:datasets}. Classification is done based on the classes defined in the second column of the table.

\begin{table}[ht!]
\centering
\caption{Datasets used in the experiments to demonstrate the error contribution and propagation methodology. These datasets are from the scientific domain.}
\begin{tabularx}{\linewidth}{ |X|X| } 
 \hline
 Dataset (notation) & Distribution of classes \\ 
 \hline
 Breast cancer (\textit{breast}) \cite{bilgin2007cell} & \textit{benign}: 151, \textit{in-situ}: 93, \textit{invasive}: 202\\
 \hline
 Brain cancer (\textit{brain}) \cite{gunduz2004cell} & \textit{glioma}: 16, \textit{healthy}: 210, \textit{inflammation}: 107\\
 \hline
  Material science 1 (\textit{matsc1}) \cite{chowdhury2016image} & \textit{dendrites}: 441, \textit{non-dendrites}: 132 \\
 \hline
 Material science 2 (\textit{matsc2}) \cite{chowdhury2016image} & \textit{transverse}: 393, \textit{longitudinal}: 48 \\
 \hline
 \end{tabularx}
\label{table:datasets}
\end{table}
These datasets have been chosen because they represent examples of real world datasets. They are noisy in the sense that they have artefacts in the images, are heavily imbalanced and are small in terms of number of samples. They are different from the very large datasets like ImageNet \cite{deng2009imagenet}, where deep learning techniques like convolutional neural networks have been shown to be superior. \cite{shin2016deep} has shown that machine learning problems involving datasets from medical imaging may be solved using pre-trained and fine-tuned neural networks rather than training them from scratch. We have therefore used pre-trained models (on the ImageNet database) such as VGGnet \cite{simonyan2014very} and InceptionNet \cite{szegedy2016rethinking} as feature extraction methods that fit naturally in the pipeline framework described here for the purpose of illustrating the error quantification methodologies.

\begin{figure*}[ht!]
\centering
\begin{subfigure}{.5\textwidth}
  \centering
  \includegraphics[scale=0.5]{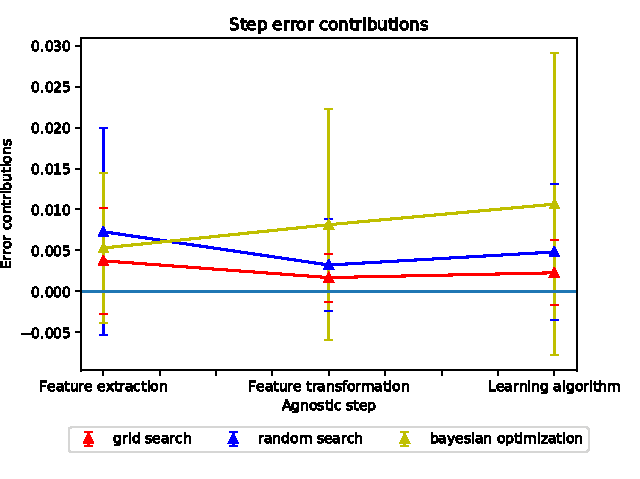}
  \caption{Contributions from steps in the pipeline}
  \label{fig:eq_steps}
\end{subfigure}%
\begin{subfigure}{.5\textwidth}
  \centering
  \includegraphics[scale=0.5]{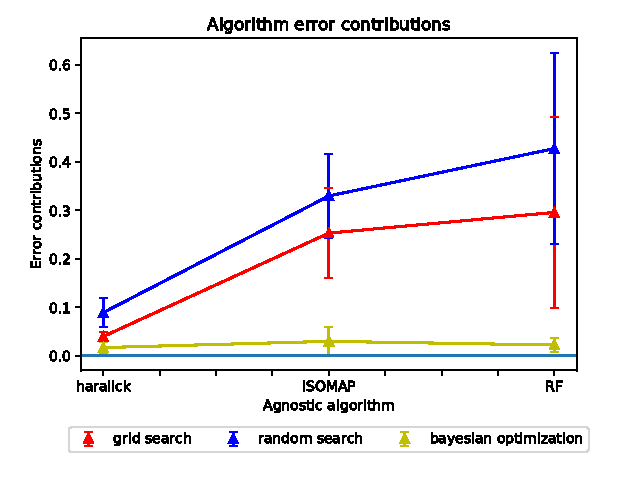}
  \caption{Contributions from algorithms in the path}
  \label{fig:eq_alg}
\end{subfigure}
\begin{subfigure}{.5\textwidth}
  \centering
  \includegraphics[scale=0.5]{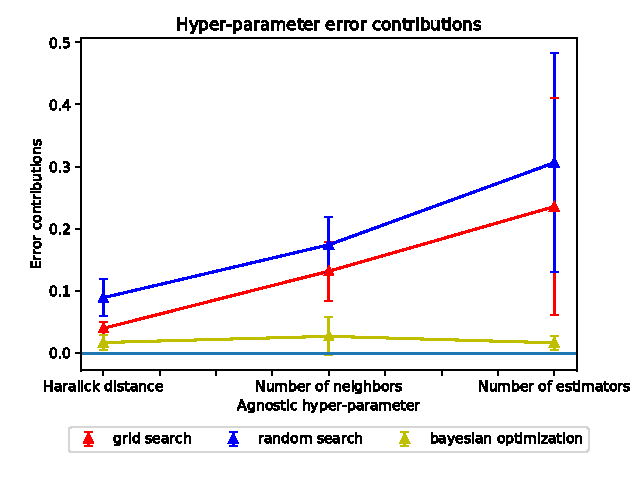}
  \caption{Contributions from hyperparameters in the path}
  \label{fig:eq_hyper}
\end{subfigure}%
\begin{subfigure}{.5\textwidth}
  \centering
  \includegraphics[scale=0.37]{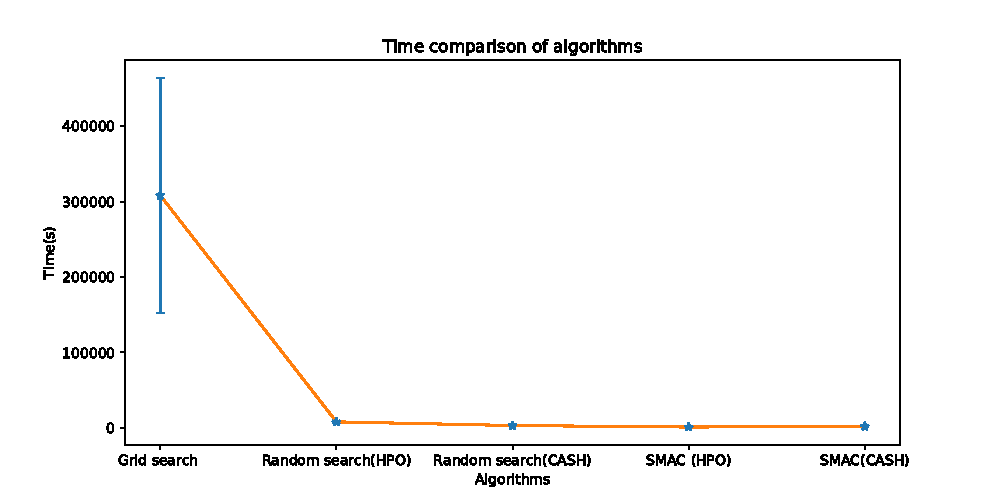}
  \caption{Comparison of computational times of optimization methods}
  \label{fig:times}
\end{subfigure}
\caption{Plots of error contributions from (\ref{fig:eq_steps}) computational steps, (\ref{fig:eq_alg}) algorithms and (\ref{fig:eq_hyper}) hyperparameters in the pipeline. Random search (blue) follows the behavior of grid search (red) more accurately than Bayesian optimization (yellow). Hence, random search maybe used to quantify the error contributions instead of grid search. In addition, we also compare the computation time for each of the algorithms of section \ref{subsec:optimization} in Fig. \ref{fig:times}. We observe that the random search and Bayesian optimization based methods (HPO and CASH) are much more efficient at computing the error contributions.}
\label{fig:EQ}
\end{figure*}

\subsection{Experimental setting}
Optimization using the 3 algorithms in section \ref{subsec:optimization} is performed using the pipeline in Fig. \ref{fig:images_pipeline} on the 4 datasets in Table \ref{table:datasets}. The domain and possible values of the hyperparameters are described in table \ref{table:hyper}. 
The convergence criteria (a hyper-hyperparameter) is set at 50 iterations of unchanging  function value for each of the optimization methods. The choice of the convergence criteria and hyperparameters are independent of the error quantification methods. Results maybe obtained by using any choice of values for these components. The continuous hyperparameters \textit{Maximum features}, \textit{C} and $\gamma$ have been discretized specifically for comparison of the optimization methods with grid search. This is because grid search requires the hyperparameters to be discretized. In general, discretization of the hyperparameters is not necessary for performing optimization.

\begin{table}[ht!]
\centering
\caption{hyperparameter domains corresponding the algorithms in Table \ref{table:algorithms_table} used by the optimization methods}
\begin{tabularx}{\linewidth}{ |X|X| }
 \hline
 hyperparameter & Values \\ 
 \hline
 \textit{Haralick distance} & [1, 2, 3, 4]\\
 \hline
 \textit{Whitening} & [True, False]\\
 \hline
  \textit{Number of neighbors} & [3, 4, 5, 6, 7] \\
 \hline
 \textit{Number of components} & [2, 3, 4]\\
 \hline
 \textit{Number of estimators} & [8,  81, 154, 227, 300]\\
 \hline
 \textit{Maximum features} & [0.3, 0.5, 0.7]\\
 \hline
 \textit{C} & [0.1, 25.075, 50.05, 75.025, 100.0] \\
 \hline
 $\gamma$ & [0.3, 0.5, 0.7] \\
 \hline
 \end{tabularx}
\label{table:hyper}
\end{table}
The error contribution and propagation values are obtained from the trials in the optimization methods described in Section \ref{sec2}. Grid search is only run once while the other algorithms are averaged over 5 runs. These results are computed on the 5-fold cross-validation error (cross-entropy loss) obtained at the end of the pipeline. Random search and Bayesian optimization (using the SMAC algorithm) are implemented on both the frameworks described in Section \ref{frameworks}. The grid search results maybe used as the gold standard to compare the performance of other optimization methods.

\subsection{Error contribution experiments}
\label{eq_expts}
Experiments based on the quantification of error contributions framework described in section \ref{EQ} are presented here. Fig. \ref{fig:EQ} shows plots of the error contribution values calculated using Eqs. \ref{eq:eq_step}, \ref{eq:eq_alg} and \ref{eq:eq_hyper} on the 4 datasets described in Table \ref{table:datasets}.

Figs. \ref{fig:eq_steps}, \ref{fig:eq_alg}, \ref{fig:eq_hyper},  show the contributions of different components (steps, algorithms and hyperparameters respectively) of the image classification pipeline.  We observe that random search is more accurate than Bayesian optimization with respect to grid search. This maybe explained by the iterative nature of the Bayesian optimization algorithm described in section \ref{bayesian}. As a result, Bayesian optimization samples fewer configurations than random search. 
Fig. \ref{fig:eq_steps} shows the results of contributions from the different computational steps in the pipeline. We observe that feature extraction has the highest contribution to the pipeline. This agrees with our intuition that feature extraction is the most important step of the pipeline. Therefore care should be taken in the choice and tuning of feature extraction algorithms. 

Fig. \ref{fig:eq_alg} shows the average of the error contributions from the algorithms over a single path. The path selected for demonstrating the error contribution from algorithms and hyperparameters experiments is \textit{Haralick texture features} - \textit{ISOMAP} - \textit{Random forests}. The choice of the path is arbitrary. This particular path is selected for the purpose of demonstration. In general, the formulation for error contributions of algorithms and hyperparameters can be used for any path in the pipeline. We observe from the results that it is more important to optimize \textit{ISOMAP} and \textit{Random forests} than \textit{Haralick texture features}. This can be explained by the fact that the number of hyperparameter configurations of \textit{Haralick texture features} is less than \textit{Random forests} and \textit{ISOMAP}. 

This can also explain the results of the contribution from hyperparameters in Fig. \ref{fig:eq_hyper}, which quantifies the contribution of hyperparameters for the same path used for the error contribution experiment for algorithms. The corresponding hyperparameters analyzed here are \textit{Haralick distance}, \textit{Number of neighbors} and \textit{Number of estimators} corresponding to the algorithms in the path \textit{Haralick texture features}, \textit{ISOMAP} and \textit{Random forests} respectively. Again, we see that the contribution  from \textit{Haralick distance}  is less than \textit{Number of estimators} and \textit{Number of neighbors}. Therefore, it is more important to tune the \textit{Number of neighbors} and \textit{Number of estimators} hyperparameters than \textit{Haralick distance}. 

In Fig. \ref{fig:times} we show the comparison of computational time to run each of the algorithms for both the HPO and CASH frameworks averaged over 4 datasets. The computational time required to run random search and Bayesian optimization are approximately the same. They are much more efficient than grid search as expected.
We observe from the results in Fig. \ref{fig:EQ} that random search maybe used to quantify the error contributions from different components more accurately than Bayesian optimization and is more efficient than grid search. Therefore, random search maybe used as a proxy for grid search for quantifying error contributions from different components of an image classification pipeline.
The individual results for each component with respect to each of the 4 datasets is shown in section 1 of the supplementary material.  

\begin{figure*}[ht!]
\centering
\begin{subfigure}{.5\textwidth}
  \centering
  \includegraphics[scale=0.5]{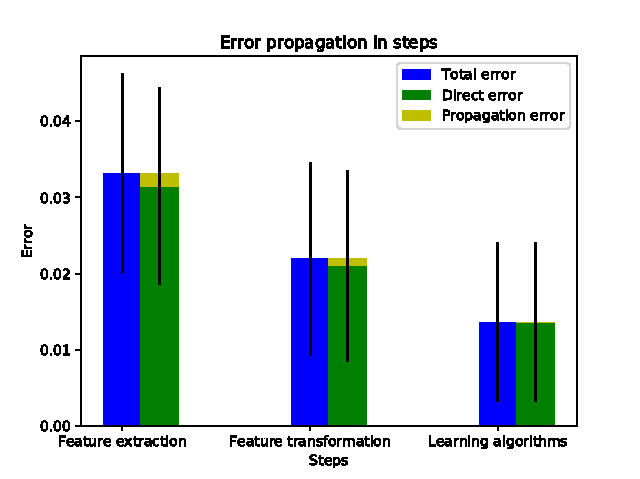}
  \caption{Error propagation from computational steps}
  \label{fig:ep_steps}
\end{subfigure}%
\begin{subfigure}{.5\textwidth}
  \centering
  \includegraphics[scale=0.5]{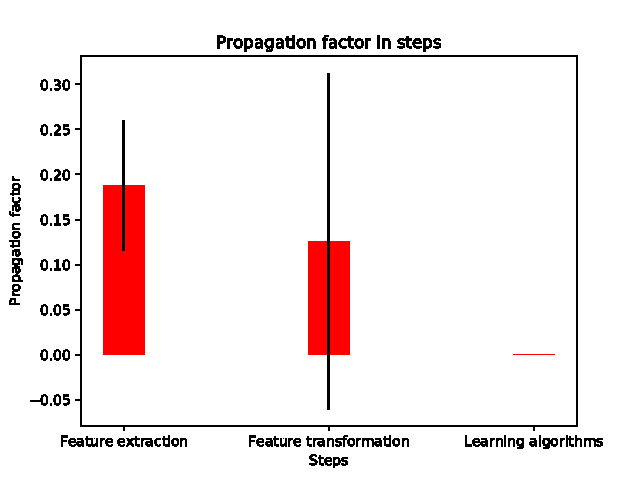}
  \caption{Propagation factor of steps in the pipeline}
  \label{fig:prop_steps}
\end{subfigure}
\begin{subfigure}{.5\textwidth}
  \centering
  \includegraphics[scale=0.5]{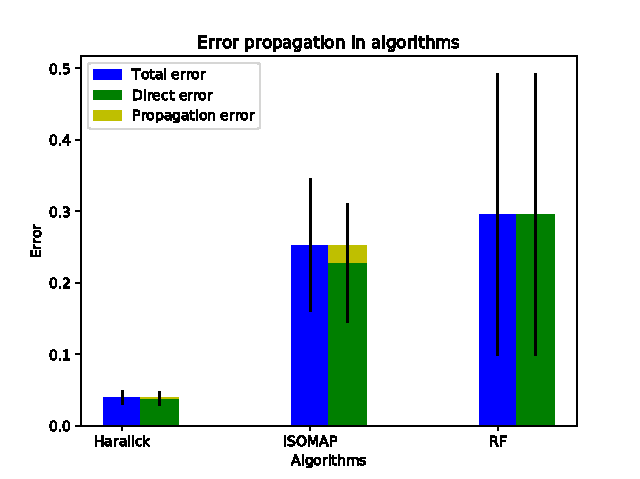}
  \caption{Error propagation from algorithms}
  \label{fig:ep_alg}
\end{subfigure}%
\begin{subfigure}{.5\textwidth}
  \centering
  \includegraphics[scale=0.5]{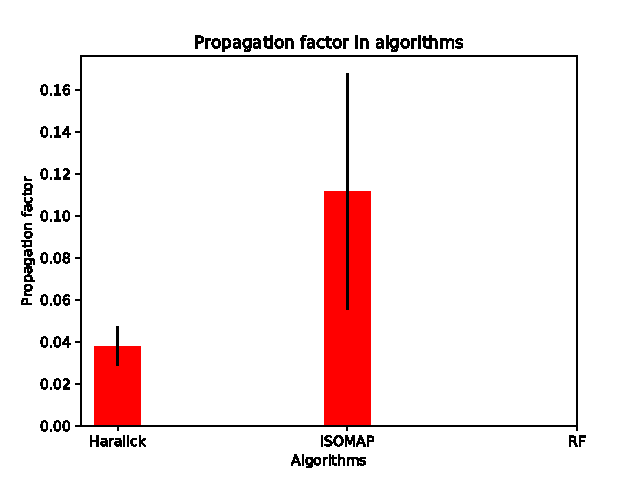}
  \caption{Propagation factor of algorithms in the path}
  \label{fig:prop_alg}
\end{subfigure}
\begin{subfigure}{.5\textwidth}
  \centering
  \includegraphics[scale=0.5]{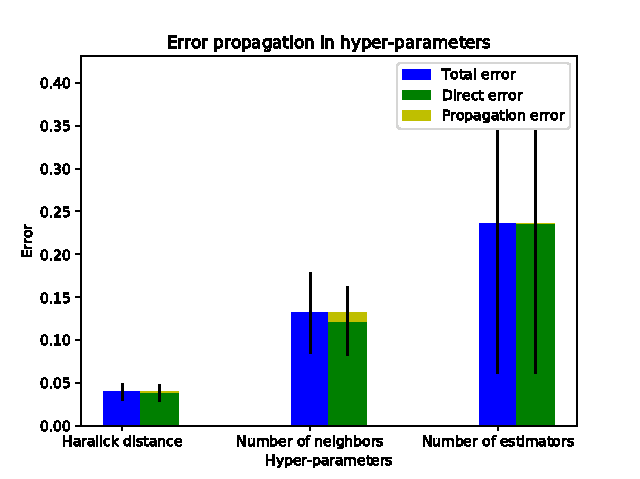}
  \caption{Error propagation from hyperparameters}
  \label{fig:ep_hyper}
\end{subfigure}%
\begin{subfigure}{.5\textwidth}
  \centering
  \includegraphics[scale=0.5]{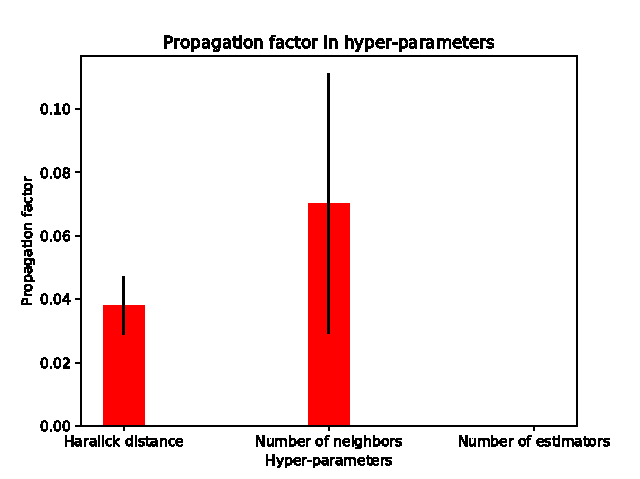}
  \caption{Propagation factor of hyperparameters in the path}
  \label{fig:prop_hyper}
\end{subfigure}

\caption{Plots of error propagation and propagation factors using the naive algorithm based methodology. We observe that the total error is dominated by the direct error.}
\label{fig:EP}
\end{figure*}

\begin{figure*}[ht!]
\centering
\begin{subfigure}{.5\textwidth}
  \centering
  \includegraphics[scale=0.5]{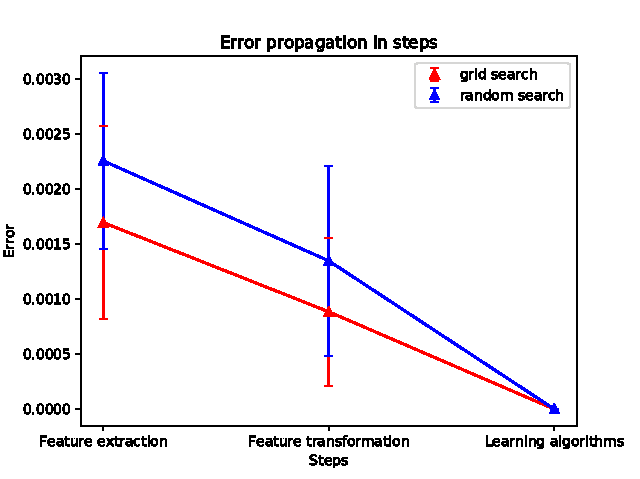}
  \caption{Comparison of error propagation on steps}
  \label{fig:steps_ep_comp}
\end{subfigure}%
\begin{subfigure}{.5\textwidth}
  \centering
  \includegraphics[scale=0.5]{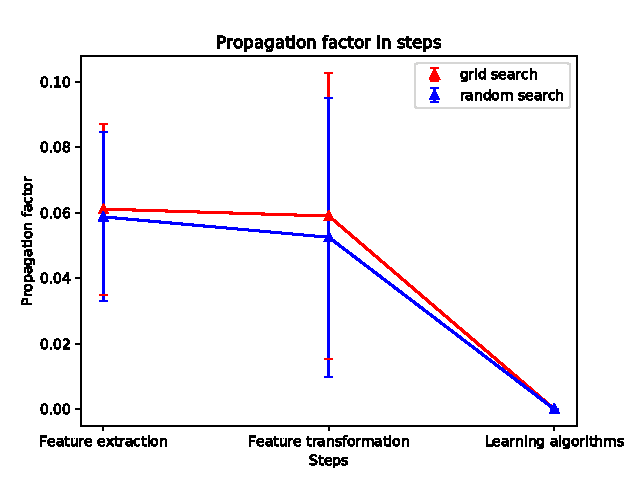}
  \caption{Comparison of propagation factor on steps}
  \label{fig:steps_prop_comp}
\end{subfigure}
\begin{subfigure}{.5\textwidth}
  \centering
  \includegraphics[scale=0.5]{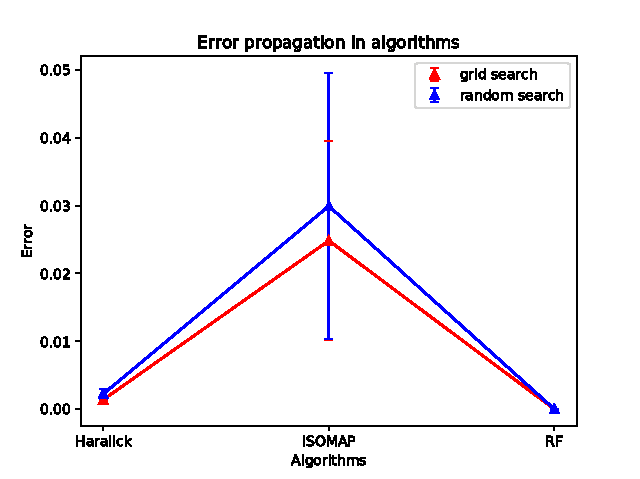}
  \caption{Comparison of error propagation on algorithms}
  \label{fig:alg_ep_comp}
\end{subfigure}%
\begin{subfigure}{.5\textwidth}
  \centering
  \includegraphics[scale=0.5]{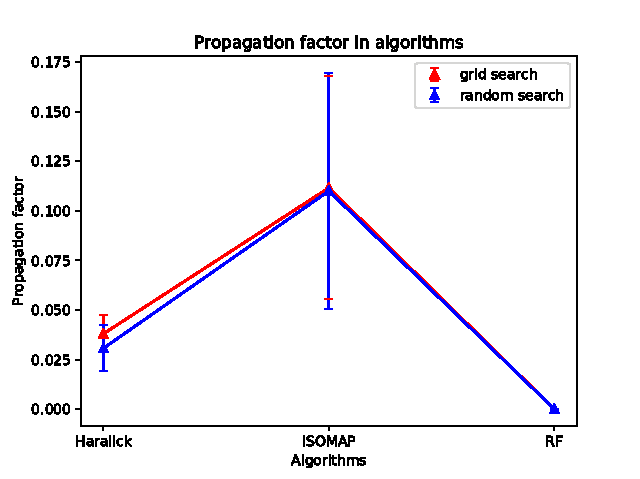}
  \caption{Comparison of propagation factor on algorithms}
  \label{fig:alg_prop_comp}
\end{subfigure}
\begin{subfigure}{.5\textwidth}
  \centering
  \includegraphics[scale=0.5]{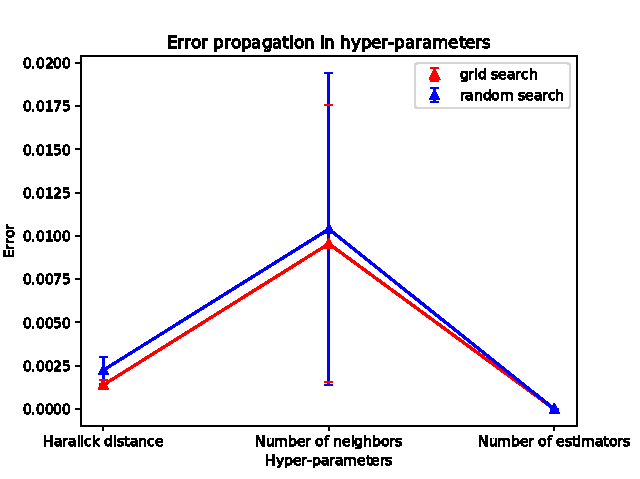}
  \caption{Comparison of error propagation on hyperparameters}
  \label{fig:hyper_ep_comp}
\end{subfigure}%
\begin{subfigure}{.5\textwidth}
  \centering
  \includegraphics[scale=0.5]{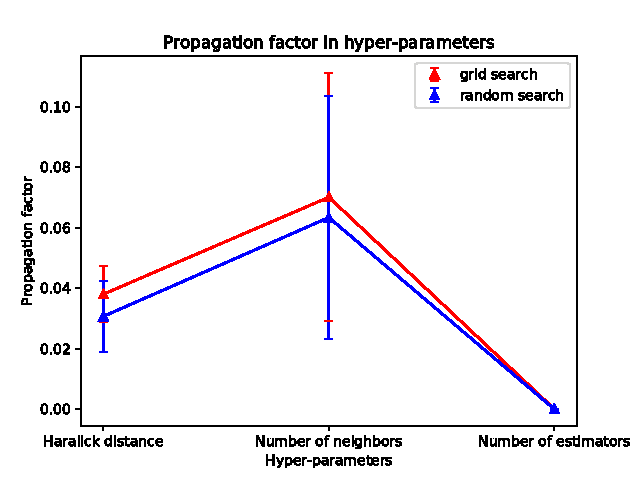}
  \caption{Comparison of propagation factor on hyperparameters}
  \label{fig:hyper_prop_comp}
\end{subfigure}

\caption{Plots of comparison error propagation and propagation factors using grid search and random search. The plots show that just like the results of error contribution in \ref{fig:EQ}, random search is able to quantify the error propagation values accurately compared to grid search.}
\label{fig:EP_comps}
\end{figure*}

\begin{figure}[H]
    \centering
    \includegraphics[scale=0.4]{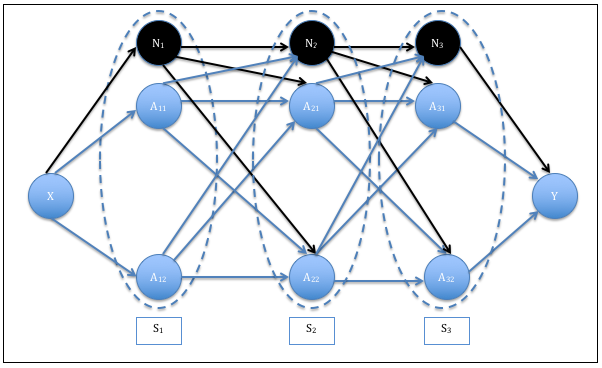}
    \caption{Pipeline including \textit{naive} algorithms (denoted in black) used in the error propagation model. }
    % The \textit{naive} algorithms serve as a bench mark for the error propagation model.}
    \label{fig:naive_pipeline}
\end{figure}

\subsection{Error propagation experiments}
\label{ep_expts}

The set of experiments in this section correspond to the error propagation model formulated in Section 3.2. The corresponding pipeline including the naive algorithms are represented in Fig. \ref{fig:naive_pipeline}.
This is an instantiation of the pipeline for the error propagation model depicted in Fig. \ref{fig:naive}.

The \textit{naive} or benchmark algorithms used in this work are pre-trained \textit{InceptionNet} \cite{szegedy2016rethinking} trained on the ImageNet \cite{deng2009imagenet} data base for feature extraction, \textit{no transformation} for feature transformation, and \textit{1-nearest neaighbors} algorithm as the classification or learning algorithm. 

These algorithms or models have been selected because they are not part of the pipeline that we are analyzing in this work. Also, these algorithms do not have any additional hyperparameters that need tuning.
In general, any set of algorithms maybe chosen to represent the \textit{naive} algorithms as long as they are not part of the original pipeline.
Fig.  \ref{fig:EP} shows the results of the error propagation framework. The results are obtained based on the grid search results and the corresponding trials. 
The first column of Fig. \ref{fig:EP} shows the error propagation from steps, algorithms and hyperparameters in figs. \ref{fig:ep_steps}, \ref{fig:ep_alg} and \ref{fig:ep_hyper} respectively . These plots show the total error $EC$, the direct error $E_{direct}$ and the propagated error $E_{propagation}$. 
These values are computed using the error propagation model defined in section \ref{subsec:naive} and averaged over the 4 datasets described in table \ref{table:datasets}.

The second column in Fig. \ref{fig:EP} shows the propagation factors $\gamma$ for the corresponding component in the first column.

The total error is same as error contribution computed for different components. As such, we observe that the direct error, propagated error and the corresponding propagation factors follow the trend of the error contribution plots in Fig. \ref{fig:EQ}. This is to be expected because from Eq. \ref{eq:EP}, we see that $\alpha$ is proportional to $E_1$ and $\gamma$ is proportional to $E_2 - E_1$ and as a result $E_{propagated}(=\gamma\alpha)$ is also proportional to the error contribution values.

We can see from the results that the total error is dominated by the direct error. The propagation error is the difference between the total error and the direct error.
We observe that in Fig. \ref{fig:prop_steps}, the propagation factor progressively reduces in magnitude from feature extraction algorithms (that are at the beginning of the pipeline) to learning algorithms (that are at the end of the pipeline). Therefore, the model captures our intuition that more error is propagated from steps that exist closer to the beginning of the pipeline. 
We do not see the same trend in algorithms and hyperparameters in figs. \ref{fig:prop_alg} and \ref{fig:prop_hyper} respectively. This is because the propagation factors is proportional to the corresponding error contribution values computed using Eqs. \ref{eq:eq_alg} and \ref{eq:eq_hyper} respectively.
We also observe that the error propagation values and propagation factor values are negligible for last step (learning algorithms) and the corresponding algorithm (RF) and hyperparameter (\textit{Number of estimators}). This is because no error is expected to be propagated from the last step of the pipeline.

The individual results of the error propagation for each component with respect to each of the 4 datasets is shown in section 2 in the supplementary material.

Fig. \ref{fig:EP_comps} shows the comparison of the error propagation and propagation factor values between grid search and random search. In section \ref{eq_expts}, we showed that random search maybe used as a proxy for grid search for computing the error contributions. We therefore only compare the results of error propagation and propagation factor between grid and random search to see if they are comparable. Bayesian optimization is not represented in the comparison because it was not able to represent the error contributions accurately according to the results in section \ref{eq_expts}.

From the results, we observe that random search follows the behavior of grid search for the error propagation values as well. This shows that random search can be used to quantify the error propagation values in addition to the error contribution values.

\section{Conclusion}
The suggested approaches involve understanding the contribution and propagation of error in data analysis pipelines. Specifically, we propose a methodology to quantify the error contributions from different parts of an image classification pipeline, namely computational steps, algorithms and hyperparameters. This methodology is described as the \textit{agnostic} method in Section \ref{EQ}. The results in Section \ref{eq_expts} show that random search is able to quantify the error contributions as well as grid search in terms of both accuracy and efficiency. The framework of Bayesian optimization is not as accurate and robust as random search due to reasons specified in section \ref{sec4}. Specifically, the sequential nature of Bayesian optimization prevents it from sampling as many configurations as random search. Hence, the error contribution estimates of Bayesian optimization are not as accurate as random search.
In general we expect optimization algorithms that have more trials in a larger region of the search space of the configurations to quantify the contributions from components in the pipeline accurately. We intend to explore the results from more hyperparameter optimization algorithms in the future like the algorithms in \cite{olson2016tpot, li2016hyperband}. 

The \textit{agnostic} methodology maybe used by machine learning practitioners to understand and interpret results of a specific machine learning problem on a particular dataset. Understanding the source of error in terms of steps, algorithms and hyperparameters will help data scientists quickly iterate over pipelines, algorithms and hyperparameters and find the best set of configurations for solving a particular task by focussing on the important components of the pipeline found from the results of error contribution. In addition, domain experts like biologists and scientists from different disciplines can use this method to understand and interpret the error originating in the pipeline to solve their specific image classification problem.

We also propose a model for error propagation to analyse the error from the pipeline even further. This is denoted as the \textit{naive} methodology, where we use \textit{naive} or benchmark algorithms in each step of the pipeline. The \textit{naive} algorithms are not part of the pipeline itself but are used as benchmarks to formulate the model. This is described in detail in section \ref{sec3}. The results from section \ref{ep_expts} show that most of the error propagates from the feature extraction algorithms followed by feature transformation algorithms. This supports our intuition that error accumulates and flows along the pipeline, with the most amount of error coming from steps at the beginning of the machine learning pipeline and it gradually reduces to zero at the end of the pipeline (learning algorithms), where all of the error is directly due to classification algorithms themselves. We also observe that the error propagation values are proportional to the error contribution values based on the designed model. This is the reason for results of error propagation from algorithms and hyperparameters as described in section \ref{ep_expts}. Fig. \ref{fig:EP_comps} shows that random search maybe used as a proxy for grid search for computing the error propagation values for a particular pipeline. This means that data scientists can quickly get accurate estimates of error contribution and propagation values from the pipeline they wish to analyze. 
In terms of future work, this formulation could be expanded to cover more data analysis problems that involve more complex algorithms. For example, this framework maybe used to interpret problems in other applications like speech recognition, text classification and even unsupervised machine learning problems. Essentially, the error quantification frameworks maybe used by any practitioner that works with pipelines for solving a machine learning problem. This framework can also be used to understand and interpret deep neural networks, which are end-to-end in nature. This maybe used for comparing the performance of candidate networks for solving the problem by quantifying the contribution and propagation of error from the hyper-parameters of the neural network.

\label{sec5}

% or


% \appendices

% \section{}

\ifCLASSOPTIONcompsoc
%   \section*{Acknowledgments}
% \else
%   % regular IEEE prefers the singular form
% \fi

% The authors would like to thank the Cognitive Immersive Systems Laboratory (CISL) at Rensselaer Polytechnic Institute headed by Dr. Hui Su for helping with the integration of this work as an application as described in section \ref{sec:application}.

\ifCLASSOPTIONcaptionsoff
  \newpage
\fi



%\bibliographystyle{IEEEtran}
\bibliographystyle{IEEEtran}
\bibliography{sample}


\begin{IEEEbiography}{Aritra Chowdhury}
Aritra Chowdhury is a Ph.D. student in the Department of Computer Science at Rensselaer Polytechnic Institute (RPI). He obtained a B.E. in Electronics and Telecommunication Engineering (2013) from Jadavpur University and an MS in Computer Science (2016) from RPI. His research interests include applications of machine learning and computer vision in medical and other scientific domains. He has a number of publications in refereed journals and conferences.
\end{IEEEbiography}

\begin{IEEEbiography}{Malik Magdon-Ismail}
Malik Magdon-Ismail is an Associate Professor of Computer Science at Rensselaer Polytechnic Institute. He obtained a BS in Physics (1993) from Yale University, an MS in Physics (1995) and a PhD in Electrical Engineering with a minor in Physics (1998) from the California Institute of Technology. He was a research fellow in the Learning Systems Group at Caltech (1998-2000) before joining Rensselaer, where he is a member of the Theory group. His research interests include the theory and applications of machine and computational learning (supervised, reinforcement and unsupervised), communication networks and computational finance. He has served on the program committees of several conferences, and is an Associate editor for Neurocomputing. He has numerous publications in refereed journals and conferences, has been a financial consultant, has collaborated with a number of companies, and has several active grants from NSF. http://www.cs.rpi.edu/∼magdon Member IEEE, Member ACM, 
\end{IEEEbiography}

%\newpage

\begin{IEEEbiography}{B{\"u}lent Yener}
B{\"u}lent Yener is a Professor in the Department of Computer Science and the founding Director of Data Science Research Center at Rensselaer Polytechnic Institute (RPI) in Troy, New York.  Dr. Yener received his MS. and Ph.D. degrees in Computer Science, both from Columbia University, in 1987 and 1994, respectively. Before joining RPI, he was a Member of the Technical Staff at the Bell Laboratories in Murray Hill, New Jersey. He is a senior member of ACM and a Fellow of IEEE (Communications Society, Computer Society, Engineering in Medicine and Biology Society).
\end{IEEEbiography}


%\vfill


\end{document}